\pdfoutput=1
\documentclass[conference]{IEEEtran}
\usepackage{amsmath,amssymb,amsfonts}
\usepackage{array}
\usepackage{booktabs}
\usepackage{graphicx}
\usepackage{tabularx}
\usepackage{float}
\usepackage{multirow}
\usepackage{subcaption}
\usepackage{url}
\graphicspath{{figures/}}
\usepackage{placeins}
\usepackage{algorithm}
\usepackage{algpseudocode}
\usepackage{algorithmicx}

\begin{document}

\title{Attention-Augmented YOLOv8 with Ghost Convolution for Real-Time Vehicle Detection in Intelligent Transportation Systems}

\author{
    Syed Sajid Ullah\textsuperscript{1,*}, 
    Muhammad Zunair Zamir\textsuperscript{2}, 
    Ahsan Ishfaq\textsuperscript{2},
    Salman Khan\textsuperscript{1}\\[4pt]
    \textsuperscript{1}\textit{School of Energy and Electrical Engineering, Chang'an University, Xi'an, China} \\
    \textsuperscript{2}\textit{School of Information Engineering, Chang'an University, Xi'an, China} \\
    \textit{*Corresponding author:} sajid@chd.edu.cn \\[2pt]
}
\maketitle

\begin{abstract}
Accurate vehicle detection is essential for autonomous driving, traffic monitoring, and intelligent transportation systems. This paper presents an enhanced YOLOv8n model that incorporates the Ghost Module, Convolutional Block Attention Module (CBAM), and Deformable Convolutional Networks v2 (DCNv2). The Ghost Module streamlines feature generation to reduce redundancy, CBAM applies channel and spatial attention to improve feature focus, and DCNv2 enables adaptability to geometric variations in vehicle shapes. These components work together to improve both accuracy and computational efficiency. Evaluated on the KITTI dataset, the proposed model achieves 95.4\% mAP@0.5---an 8.97\% gain over standard YOLOv8n---along with 96.2\% precision, 93.7\% recall, and a 94.93\% F1-score. Comparative analysis with seven state-of-the-art detectors demonstrates consistent superiority in key performance metrics. An ablation study is also conducted to quantify the individual and combined contributions of Ghost Module, CBAM, and DCNv2, highlighting their effectiveness in improving detection performance. By addressing feature redundancy, attention refinement, and spatial adaptability, the proposed model offers a robust and scalable solution for vehicle detection across diverse traffic scenarios.
\end{abstract}



\section{Introduction}
Vehicle detection has become increasingly critical in various domains, including autonomous driving, traffic monitoring, and intelligent transportation systems (ITS). The ability to accurately identify vehicles in diverse and complex environments is essential for ensuring safety, optimizing traffic flow, and enhancing the overall efficiency of transportation networks. As urbanization accelerates and vehicle populations grow, the challenges associated with vehicle detection, such as occlusion, varying scales, and diverse orientations, have intensified. Consequently, there is a pressing need for advanced detection algorithms that can effectively address these challenges while maintaining computational efficiency \cite{Yan2021}.

\begin{table*}[t]
    \centering
    \caption{Evolution of YOLO Models}
    \begin{tabularx}{\textwidth}{@{}p{1.5cm} p{2.5cm} p{1cm} p{12cm}@{}} 
        \toprule
        \textbf{Ref.} & \textbf{YOLO Version} & \textbf{Year} & \textbf{Main Advancements} \\ 
        \midrule
        \cite{redmon2016you}         & YOLOv1   & 2015 & Real-time detection using a grid-based unified model for detection and regression. \\ 
        \cite{redmon2017yolo9000}    & YOLOv2   & 2016 & Added anchor boxes, dimension clustering, and multi-scale training for small object detection. \\ 
        \cite{redmon2018yolov3}      & YOLOv3   & 2018 & Introduced Darknet-53 backbone, multi-label classification, and a three-scale feature pyramid. \\ 
        \cite{bochkovskiy2020yolov4} & YOLOv4   & 2020 & Incorporated Mosaic augmentation, CSPDarknet53 backbone, and PAN/SAM neck improvements. \\ 
        \cite{jocher2020ultralytics} & YOLOv5   & 2020 & PyTorch-native version with autoanchor, hyperparameter evolution, and simplified architecture. \\ 
        \cite{keylabs2023}           & \textbf{YOLOv8} & \textbf{2023} & Used dynamic convolution, an anchor-free split head, and trained on multiple datasets. \\ 
        \bottomrule
    \end{tabularx}
    \label{tab:yolo_evolution}
\end{table*}

Real-time and accurate vehicle detection remains fundamental to modern intelligent transportation systems. While deep learning-based object detection models have shown remarkable progress, they still face challenges in detecting small, occluded vehicles in complex urban scenarios. Additionally, computational efficiency remains a crucial factor for real-world deployment, particularly in edge devices and autonomous vehicles with limited resources.

The YOLO (You Only Look Once) family of detectors has emerged as a leading framework for real-time object detection due to its balance of speed and accuracy. The evolution of YOLO models are demonstrated in Table~\ref{tab:yolo_evolution}. YOLOv8, the latest iteration, incorporates several advancements over its predecessors, including improved backbone networks and enhanced feature extraction mechanisms. However, despite its strengths, YOLOv8 still faces limitations in detecting small and occluded objects, which are prevalent in real-world scenarios \cite{Ding2024}. While multiple variants of YOLOv8 exist, our work utilizes YOLOv8n due to its lightweight architecture and superior efficiency, making it suitable for deployment in resource-constrained environments.

This study aims to address these limitations by integrating three innovative components: the Ghost Module, the Convolutional Block Attention Module (CBAM), and Deformable Convolutional Networks v2 (DCNv2) \cite{Tang2022,Chien2024,Li2025}. The Ghost Module is designed to improve feature extraction by reducing redundancy in the feature maps, thereby allowing the model to focus on more relevant features \cite{Tang2022}. CBAM further enhances this capability by applying attention mechanisms that enable the model to prioritize important features across multiple scales, which is particularly beneficial in complex environments where vehicles may be partially obscured or vary significantly in size \cite{Chien2024}. DCNv2 introduces learnable offsets and modulation scalars into convolutional kernels, improving the model’s robustness to geometric transformations in object appearance \cite{Li2025}. Additionally, the integration of DCNv2 increases the model's robustness to variations in object position, scale, and occlusion, which are common challenges in vehicle detection tasks \cite{Guo2022}.

The main contributions of this paper are:
\begin{itemize}
    \item A novel YOLOv8-based architecture that strategically integrates Ghost Module, CBAM, and DCNv2 to enhance vehicle detection performance.
    \item A comprehensive evaluation of the proposed architecture on standard vehicle detection benchmarks, demonstrating significant improvements in accuracy and computational efficiency.
    \item An ablation study that quantifies the individual and combined contributions of Ghost Module, CBAM, and DCNv2, validating their impact on overall detection performance.
    \item Empirical evidence showing that the proposed model significantly outperforms the baseline YOLOv8.
    
\end{itemize}

The structure of this paper is as follows: Section 2 reviews the related work on vehicle detection methods. Section 3 presents the methodology, including the proposed YOLOv8-based model, the integration of Ghost Module, CBAM, and DCNv2, as well as the dataset and evaluation metrics. Section 4 presents a comparative analysis of the proposed model with baseline and state-of-the-art methods, followed by a discussion of the results, highlighting key improvements and their implications. Section 5 concludes the study and suggests future research directions.

Vehicle detection has garnered significant attention in recent years, particularly with the advent of deep learning techniques that have revolutionized the field of computer vision. The YOLO family of models has established itself as a leading framework for real-time object detection, with each iteration introducing significant improvements. The evolution from YOLOv1 to YOLOv8 has been extensively documented \cite{terven2023}, with initial versions introducing grid-based detection and anchor boxes, while later iterations like YOLOv4 and YOLOv5 brought enhanced performance through optimizations including advanced data augmentation techniques and more flexible architectures. YOLOv8, introduced in 2023, represents the current state-of-the-art, featuring significant architectural improvements, including a more efficient backbone, enhanced feature extraction, and better accuracy-speed trade-offs \cite{safaldin2024}.

Numerous studies have explored various methodologies to enhance vehicle detection accuracy and efficiency. Yang et al. proposed an improved vehicle detection system that integrates double-layer Long Short-Term Memory (LSTM) modules, demonstrating enhanced performance on the KITTI dataset and a self-built dataset collected from Taiwanese highways \cite{yang2022}. In tunnel environments, Kim's research revealed the limitations of traditional vehicle detection methods, such as the Aggregate Channel Features (ACF) and Fast R-CNN, which struggled in conditions with reduced visibility \cite{kim2020}. Chavan's work on YOLOv5 illustrates the versatility of deep learning models in vehicle detection, emphasizing the importance of feature engineering and the sliding-window technique \cite{chavan2023}. Moreover, Jiang's study on multi-target detection for roads utilized manually designed features combined with classifiers like SVM and AdaBoost \cite{jiang2024}, while Xu introduced a lightweight vehicle detection network based on YOLOv5 for intelligent transportation applications \cite{xu2023}.

The integration of attention mechanisms has been a focal point in recent research for enhancing object detection performance. Lee et al. proposed a convolutional neural network that employs selective multi-stage feature fusion to enhance vehicle detection performance \cite{lee2018}. CBAM, proposed by \cite{Chien2024}, has emerged as a particularly effective attention mechanism that sequentially applies channel and spatial attention to refine feature maps. Several studies have demonstrated the effectiveness of CBAM in object detection tasks, with \cite{Ma2024} reporting significant improvements in detection accuracy, particularly for small objects, when integrating CBAM into YOLOv5. Reference \cite{li2020} combined CBAM with feature pyramid networks for enhanced object detection in aerial imagery. Furthermore, reference \cite{an2022} addressed the challenge of occlusion in vehicle detection by incorporating countermeasure learning into the RCNN framework, thereby improving detection accuracy in cluttered environments.

The Ghost Module, introduced by \cite{Tang2022}, addresses the computational inefficiency of traditional convolutional networks by generating more feature maps with fewer parameters. This approach significantly reduces the computational cost and model size without substantial performance degradation by applying standard convolution to generate a small set of intrinsic feature maps, followed by cheap linear operations to generate more feature maps. Similarly, Deformable Convolutional Networks v2 (DCNv2), proposed by \cite{zhu2019dcnv2}, enhance the original DCN by introducing a modulation mechanism that adjusts the influence of each sampling point. This enables more precise control over the deformation process and improved feature extraction for objects with complex shapes.

\begin{table*}[htbp]
  \centering
  \caption{Literature Review: Recent Detection Methods on KITTI}
  \label{tab:results_table2}
  \begin{tabularx}{\textwidth}{@{}%
      >{\raggedright\arraybackslash}p{1.0cm}  
      >{\raggedright\arraybackslash}p{3.8cm}  
      >{\raggedright\arraybackslash}p{2.2cm}  
      >{\raggedright\arraybackslash}p{9.8cm}  
    @{}}
    \toprule
    \textbf{Reference} & \textbf{Method} & \textbf{Dataset(s)} & \textbf{Key Insights} \\
    \midrule

    \cite{behera2024} &
    Pseudo-RGB BEV to YOLOv8 &
    KITTI, TIAND &
    LiDAR to BEV-RGB conversion achieves 86.3\% mAP with real-time inference at 28.3 FPS. \\

    \cite{Cong2024} &
    YRDM: Efficient mining + dynamic confidence &
    KITTI &
    15\% reduction in parameters while maintaining 80\% mAP; achieves 35 FPS on RTX-3090. \\

    \cite{peng2024} &
    YOLOv8 + CBAM + multi-scale fusion &
    KITTI &
    Improves small-object recall by 4\% over baseline YOLOv8 with 10\% latency increase. \\

    \cite{safaldin2024} &
    Motion-aware YOLOv8 &
    KITTI, LASIESTA &
    Preprocessing improves moving-object precision by 6\%; struggles with static targets in cluttered scenes. \\

    \cite{shen2024} &
    YOLOv5s + pruning and quantization &
    KITTI &
    Model size reduced by 3× while maintaining 75\% mAP (12\% drop in low-light conditions). \\

    \cite{Ye2023} &
    GBForkDet: Ghost module + Bi-OD Neck &
    KITTI &
    Ghost modules reduce parameters by 20\%; redesigned neck boosts mAP on occlusions by 5\%. \\

    \cite{zhao2024z} &
    Z-YOLOv8s: SSD-style head + WIoU loss &
    KITTI &
    SSD-style head increases mAP by 3\% in dense urban scenes; highway performance not reported. \\
    \bottomrule
  \end{tabularx}
\end{table*}

Recent advancements have also seen the emergence of hybrid approaches that combine different sensing modalities. Reference \cite{wang2019} proposed a real-time vehicle detection algorithm that fuses vision and LiDAR point cloud data, effectively leveraging the strengths of both modalities to enhance detection accuracy, particularly for small and occluded targets. Also, Table~\ref{tab:results_table2}
 is showing brief Summary of Related Works including Datasets Used, and Key Contributions. The landscape of vehicle detection research is characterized by a continuous pursuit of improved accuracy and efficiency through innovative methodologies, including deep learning architectures, attention mechanisms, and multi-sensor fusion techniques.

Despite the advancements in vehicle detection models, existing architectures often struggle with computational efficiency and robustness in complex environments. Traditional YOLO-based models, including the baseline YOLOv8, exhibit limitations in accurately detecting small or partially occluded vehicles due to insufficient feature extraction and poor adaptation to geometric variations. Additionally, standard convolutional modules in these models tend to generate redundant features, which not only increase computational cost but also reduce the model’s efficiency. Moreover, the lack of attention mechanisms often leads to decreased performance in scenarios where vehicles vary significantly in size or are surrounded by cluttered backgrounds. These challenges limit the practical application of such models in real-time and dynamic settings.

To address these shortcomings, the proposed model strategically integrates the Ghost Module to reduce redundancy, CBAM to enhance feature discrimination through attention mechanisms, and DCNv2 to handle geometric transformations. These modifications collectively improve detection accuracy, robustness, and computational efficiency, making the proposed model more suitable for real-world intelligent transportation systems.


\section{Methodology}\label{sec3}

\subsection{YOLOv8 Architecture}

The YOLOv8 architecture represents a state-of-the-art single-stage object detection framework characterized by its efficient and robust design. Building upon the strengths of previous YOLO versions, YOLOv8 introduces several novel improvements aimed at enhancing both performance and computational efficiency. The architecture is composed of three primary components: the Backbone Network (CSPDarknet), the Neck (Feature Pyramid Network), and the Detection Head.

\subsubsection{Backbone Network (CSPDarknet)}

The Backbone Network in YOLOv8 is built upon CSPDarknet, which utilizes Cross-Stage Partial (CSP) connections to enhance feature extraction while maintaining computational efficiency. The CSP connections split the feature maps into two parts and process them separately, thereby reducing computational cost without sacrificing feature quality. This approach is particularly beneficial for maintaining strong gradient flow during training, leading to faster convergence and better performance.

CSPDarknet incorporates a series of convolutional layers with varying kernel sizes, allowing the network to learn features at multiple scales. This design is crucial for detecting objects of different sizes, enabling the model to handle both small and large objects effectively. The architecture also integrates residual blocks and dilated convolutions, which help in capturing fine-grained details while ensuring minimal computational overhead. These elements make CSPDarknet both lightweight and powerful, ensuring robust feature extraction.

Additionally, YOLOv8 uses advanced activation functions such as Mish or SiLU. These activation functions have been shown to improve model convergence and performance, offering smoother gradients during backpropagation and better generalization. Batch normalization is applied after each convolutional operation, stabilizing the learning process and accelerating convergence by reducing internal covariate shift. This combination of features allows YOLOv8 to effectively learn and generalize complex patterns from the input data.

\subsubsection{Neck (Feature Pyramid Network)}

The Neck of YOLOv8 is designed to aggregate and refine features from multiple scales, making the model capable of detecting objects at various levels of abstraction. The Feature Pyramid Network (FPN) is used to combine low-level, high-resolution features with high-level, semantically rich features. This multi-scale feature fusion enables the model to improve detection performance, especially for objects that appear at different sizes within the image.

The network also employs the Path Aggregation Network (PAN), which enhances the feature fusion process by improving the flow of information across different stages. Unlike traditional FPNs, PAN introduces both top-down and bottom-up pathways, allowing the network to incorporate fine-grained low-level features alongside high-level context. This dual information flow ensures that the model is capable of detecting small objects in the foreground while maintaining the ability to recognize large objects in the background.

In addition to PAN, YOLOv8 employs cross-stage feature fusion and adaptive feature selection mechanisms to dynamically prioritize features based on the input data. These methods help the network to focus on the most relevant features for object detection tasks, leading to improved accuracy and efficiency. By leveraging multiple feature scales and refining them through adaptive fusion strategies, the Neck component ensures that the network can detect objects of varying sizes and complexities effectively.

\subsubsection{Detection Head}

The Detection Head in YOLOv8 is designed to generate accurate object predictions. One of the key innovations in YOLOv8 is its adoption of an anchor-free detection strategy. Unlike traditional object detection models, which rely on predefined anchor boxes, YOLOv8 directly predicts the bounding box coordinates relative to the grid cells. This eliminates the need for anchor box generation and matching, simplifying the detection process and making the model faster and more flexible.

The Detection Head is also decoupled into two distinct branches: one for classification and one for bounding box localization. This decoupling allows each task to be handled independently, reducing the competition between classification and localization tasks that often occurs in traditional models. As a result, the model can focus more on each task, leading to more accurate predictions. The decoupling also helps reduce the complexity of the model, allowing it to run more efficiently without sacrificing performance.

In addition to bounding box coordinates and class probabilities, the Detection Head also computes confidence scores, which indicate the model’s certainty about its predictions. This is crucial for object detection, as it helps in filtering out low-confidence detections that are likely false positives. The combination of anchor-free detection and a decoupled head contributes significantly to the overall speed and accuracy of YOLOv8, making it suitable for a variety of real-time applications.

\subsubsection{Computational Efficiency and Speed}

YOLOv8 is designed to be computationally efficient while maintaining high accuracy, making it ideal for real-time applications. The use of CSP connections in the Backbone Network reduces the computational cost without compromising the quality of feature extraction. Additionally, the decoupled Detection Head and anchor-free detection strategy simplify the model’s architecture, resulting in faster inference speeds. These design choices enable YOLOv8 to achieve high performance on a wide range of hardware, from GPUs to edge devices.

The network also employs several performance optimization techniques. Quantization and pruning are used to reduce the model’s size and computational requirements, making it more suitable for deployment on resource-constrained devices. Quantization reduces the precision of the model's weights, while pruning removes redundant neurons and weights, resulting in a smaller and more efficient model. Furthermore, gradient checkpointing is employed to minimize memory usage during training, allowing the model to scale more effectively and be trained on deeper architectures.

Real-time inference is achieved through a combination of reduced floating-point operations (FLOPs), hardware acceleration, and optimized layer execution. These optimizations ensure that YOLOv8 can perform object detection tasks at high speed, even in demanding applications such as autonomous driving, video surveillance, and robotics. By optimizing both the architecture and the computational processes, YOLOv8 strikes a perfect balance between accuracy, speed, and resource efficiency.

\subsubsection{Performance Optimization}

In addition to the structural optimizations mentioned above, YOLOv8 includes several performance optimization strategies to further enhance its efficiency. Techniques like model quantization and pruning not only reduce the model size but also decrease the number of operations required for inference, leading to faster processing times and lower memory usage. These optimizations make YOLOv8 suitable for real-time deployment on devices with limited computational power, such as mobile phones, drones, and embedded systems.

The use of gradient checkpointing during training helps manage memory consumption by recomputing intermediate results instead of storing them, allowing for more efficient use of available resources. This is particularly important for training larger models that require substantial memory. These optimizations, combined with the efficient design of the YOLOv8 architecture, ensure that the model can achieve high detection accuracy while running efficiently in real-time environments.

Finally, YOLOv8 has been designed to perform well across various hardware platforms. The architecture is compatible with GPUs for high-speed training and inference, while the optimized model can also be deployed on edge devices such as mobile phones or embedded systems without significant loss in performance. This flexibility makes YOLOv8 a versatile choice for a wide range of object detection tasks, from autonomous vehicles to security systems.

\subsection{Proposed Modifications}

\subsubsection{Ghost Module Integration}
To reduce computational redundancy and model complexity, we integrate the Ghost Module \cite{han2020ghostnet} into the YOLOv8 architecture. Instead of generating all feature maps via standard convolution, Ghost Modules produce a small set of intrinsic feature maps followed by additional maps generated through cheap linear operations (e.g., depthwise convolutions). Mathematically:

\begin{align}
X &= W * I \\
Y_i &= \Phi_i(X), \quad i = 1, 2, \ldots, s \\
Y &= \text{Concat}(X, Y_1, Y_2, \ldots, Y_s)
\end{align}

Here, $I$ is the input, $W$ is the convolution kernel, $\Phi_i$ represents linear transformations, and $s$ is the number of ghost features per intrinsic map.

\begin{figure}[h]
    \centering
    \includegraphics[width=0.8\linewidth]{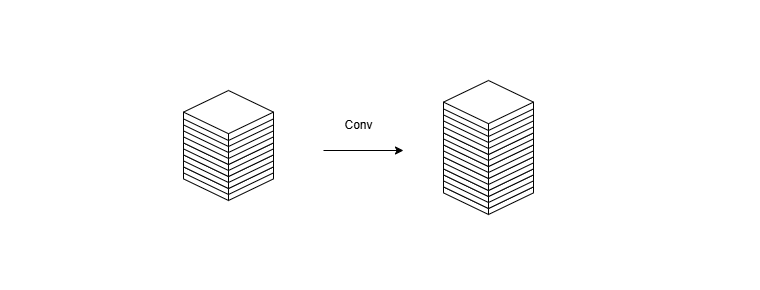}
    \caption{Convolution Layer}
    \label{fig:conv-layer}
\end{figure}

\begin{figure}[h]
    \centering
    \includegraphics[width=0.8\linewidth]{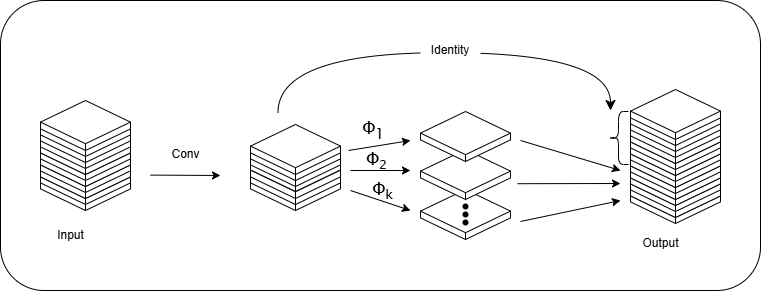}
    \caption{Ghost Module}
    \label{fig:ghost-module}
\end{figure}

The Ghost Module is deployed in the neck of YOLOv8, replacing standard convolution blocks while preserving the network's feature hierarchy. This modification enhances feature diversity and extraction efficiency with minimal computational overhead, enabling real-time inference on resource-constrained devices. The convolutional layer and architecture of Ghost Module is shown in Figure~\ref{fig:conv-layer} and Figure~\ref{fig:ghost-module}, respectively.

\vspace{1em} 

\subsubsection{Convolutional Block Attention Module (CBAM)}

The Convolutional Block Attention Module (CBAM) \cite{woo2018cbam} enhances feature representations by applying attention mechanisms in both channel and spatial dimensions. It consists of two sequential submodules: the Channel Attention Module (CAM) and the Spatial Attention Module (SAM).

\textbf{Channel Attention:}
Given an input feature map $F \in \mathbb{R}^{C \times H \times W}$, average-pooling and max-pooling are applied spatially:
\begin{equation}
F_{avg}^c = \text{AvgPool}(F), \quad F_{max}^c = \text{MaxPool}(F)
\end{equation}
These are passed through a shared MLP with weights $W_0$, $W_1$:
\begin{equation}
M_c(F) = \sigma(\text{MLP}(F_{avg}^c) + \text{MLP}(F_{max}^c))
\end{equation}
The refined feature is obtained by:
\begin{equation}
F' = M_c(F) \otimes F
\end{equation}

\textbf{Spatial Attention:}
From $F'$, spatial descriptors are computed:
\begin{equation}
F_{avg}^s = \text{AvgPool}(F'), \quad F_{max}^s = \text{MaxPool}(F')
\end{equation}
These are concatenated and passed through a convolution layer:
\begin{equation}
M_s(F') = \sigma(f^{7 \times 7}([F_{avg}^s; F_{max}^s]))
\end{equation}
The final output is:
\begin{equation}
F'' = M_s(F') \otimes F'
\end{equation}

CBAM enhances the model’s capacity to focus on informative features and suppress irrelevant background noise, leading to improved accuracy and robustness in tasks such as object detection, especially under occlusions or cluttered scenes.

\subsubsection{Deformable Convolutional Networks v2 (DCNv2)}

Deformable Convolutional Networks v2 (DCNv2) \cite{zhu2019dcnv2} extend the standard convolution operator by learning both spatial offsets and modulation scalars for each sampling location, thereby enabling adaptive reshaping of the receptive field to better capture geometric transformations and object deformations.  In DCNv2, separate convolutional layers predict a set of offsets \(\{\Delta p_k\}\) and modulation factors \(\{\Delta m_k\}\), which are applied to the regular grid positions \(\{p_k\}\) of the convolutional kernel.  The output at location \(p_0\) is computed as  

\begin{equation}
y(p_0) = \sum_{k=1}^{K} w(k)\,\bigl[x\bigl(p_0 + p_k + \Delta p_k\bigr)\bigr]\,\Delta m_k,
\label{eq:dcnv2}
\end{equation}
 
where \(w(k)\) denotes the weight of the \(k\)-th kernel element and \(x(\cdot)\) denotes the input feature (with bilinear interpolation for non-integer locations).  By jointly optimizing \(\Delta p_k\) and \(\Delta m_k\) along with the convolutional weights, DCNv2 dynamically adjusts both where and how much to sample, which substantially improves the modeling of objects under scale variation, occlusion, and non-rigid deformation.  When integrated into the detection head of YOLOv8, these learned geometric adaptations result in more precise feature extraction and stronger localization performance, particularly in challenging scenarios involving irregular object shapes and poses.

\begin{figure}
    \centering
    \includegraphics[width=0.9\linewidth]{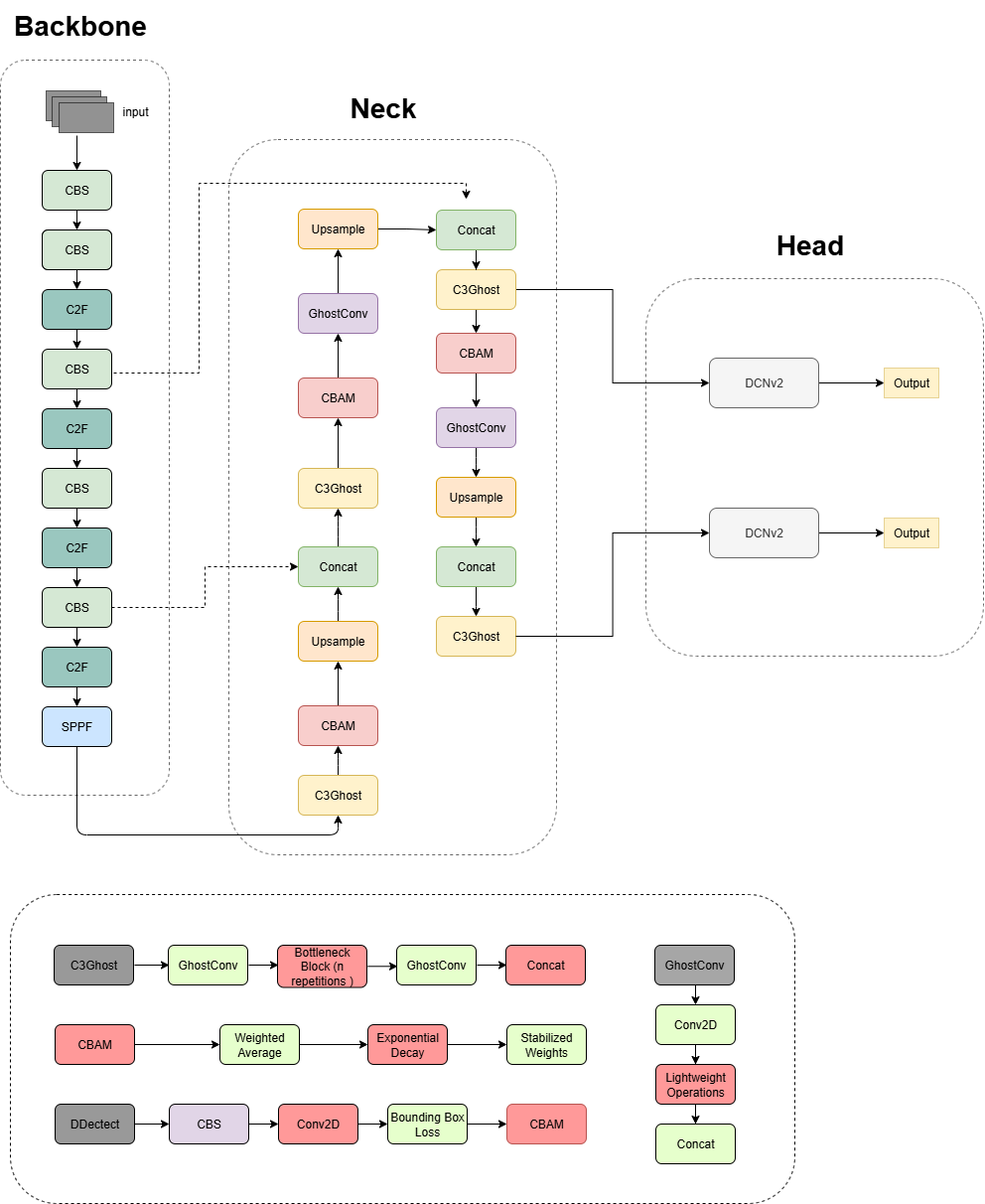}
    \caption{Architecture of the Enhanced YOLOv8 Framework Integrating Ghost Modules, CBAM, and DCNv2}
    \label{fig:figure 3}
\end{figure}


\begin{algorithm}[htbp]
\caption{Enhanced YOLOv8 with Ghost, CBAM, and DCNv2}
\label{alg:proposed}
\begin{algorithmic}[1] 
\State \textbf{Input:} Training images $I$, ground truth boxes $G$
\State Apply mosaic data augmentation on KITTI dataset
\State Integrate Ghost Module into YOLOv8 neck for efficient feature extraction
\State Insert CBAM in backbone and neck for attention-based refinement
\State Replace standard convolutions in head with DCNv2

\For{each epoch $t = 1$ to $T$}
\For{each batch of images}
\State Extract multi-scale features $F$ using modified backbone
\State Refine $F$ using CBAM and Ghost Module $\rightarrow F'$
\State Predict bounding boxes $B$ using DCNv2 head
\For{each predicted box $b \in B$}
\State Compute IoU between $b$ and ground truth $G$
\State Calculate loss $L$ (classification, localization, IoU terms)
\EndFor
\State Backpropagate loss and update model weights
\EndFor
\State Evaluate precision, recall, and time on validation set
\EndFor
\State \textbf{Output:} Trained YOLOv8-based detection model
\end{algorithmic}
\end{algorithm}

\subsubsection{Proposed Model Architecture}

The proposed model builds upon the YOLOv8n architecture and introduces key architectural modifications to enhance detection performance while maintaining computational efficiency. This section details the structure and functionality of each component in the modified network, including the \textbf{Backbone}, \textbf{Neck}, and \textbf{Head}, along with the custom modules incorporated such as GhostConv, CBAM, and DCNv2. An overview of the proposed architecture is presented in Figure~\ref{fig:figure 3}.

The backbone of the model is designed to extract hierarchical visual features from the input image using a series of convolutional and residual connections. Initially, the input image is passed through multiple CBS blocks, each comprising a convolutional layer followed by batch normalization and the SiLU activation function. These layers enable early-stage feature extraction and downsampling.

Subsequently, the model incorporates several C2F (Cross Stage Partial Fusion) modules, which are lightweight residual blocks that promote efficient feature reuse and better gradient flow across layers. This architecture is well-suited for deep but lightweight models, as it reduces the number of parameters while retaining expressive power.

To capture multi-scale spatial information, a Spatial Pyramid Pooling Fast (SPPF) module is used at the end of the backbone. This component aggregates features across different receptive fields, improving robustness to object scale variations, which is particularly beneficial in vehicle detection where size diversity is common.

The neck is responsible for aggregating features from different spatial resolutions and enhancing their discriminative capacity. In this design, the neck includes multiple \texttt{C3Ghost} modules—adapted from YOLO’s C3 block but built with GhostConv layers for improved efficiency.

At each level of the neck, a \texttt{CBAM} (Convolutional Block Attention Module) is inserted to refine the extracted features by applying both channel-wise and spatial attention. This attention mechanism allows the model to selectively emphasize salient regions such as vehicle contours, lights, and structural parts, thus improving detection robustness in cluttered scenes.

The upsampling operations in the neck are followed by concatenation with feature maps from earlier stages (skip connections), enabling the integration of both low-level and high-level contextual information. These fused features are subsequently processed by GhostConv layers to retain lightweight processing while preserving informative representations.

The detection head is designed to generate final predictions for object location and classification. At each output scale, a \texttt{DCNv2} (Deformable Convolution v2) layer is employed. Unlike standard convolutions, DCNv2 dynamically adjusts the receptive field by learning spatial offsets, making it particularly effective at modeling geometric variations such as rotated or occluded vehicles.

Each DCNv2 output is passed through a $1\times1$ convolution layer to produce the final prediction tensor, which includes bounding box coordinates, object confidence scores, and class probabilities. The prediction tensor follows the format $[B, 3, H, W, 85]$, where 85 represents 4 box coordinates, 1 objectness score, and 80 class probabilities, consistent with the COCO detection standard. The overall training and inference procedure for the enhanced YOLOv8 model is outlined in \textbf{Algorithm~1}.

The integration of GhostConv, CBAM, and DCNv2 contributes synergistically to the proposed model’s performance. GhostConv significantly reduces the number of floating point operations (FLOPs) by generating feature maps through inexpensive transformations. CBAM modules help the model to concentrate on the most relevant spatial and channel-wise regions, thereby improving focus on key vehicle parts. DCNv2 further enhances spatial adaptability, enabling precise localization of vehicles under non-ideal conditions such as skewed angles and occlusions.

\subsection{Dataset and Preprocessing}

\subsubsection{Data Composition}

The dataset used for training and evaluating the proposed model is the KITTI dataset, a well-established benchmark in the field of object detection for autonomous driving. KITTI offers approximately 70,000 real-world images collected from diverse driving environments, including urban streets, highways, and rural roads. This variety enables robust model training by exposing it to different vehicle types, object scales, traffic densities, and scene complexities. Urban scenarios in the dataset often feature occlusions and dynamic traffic, while highway scenes provide more structured, high-speed conditions with fewer obstacles. The presence of vehicles at varying distances also allows the model to learn across multiple object scales, enhancing its generalization capability.

While KITTI is limited in terms of extreme lighting variations and adverse weather conditions, as it was mostly captured in clear daytime environments, it remains a strong foundation for evaluating models under realistic, everyday driving scenarios. The dataset includes several vehicle classes such as cars, vans, trucks, and trams, supporting a wide range of detection tasks relevant to intelligent transportation systems.

\subsubsection{Preprocessing Techniques}

The preprocessing pipeline applies several critical techniques to ensure that the data is properly prepared for training and that the model can generalize well to unseen data. Various steps are employed to improve the robustness and consistency of the input data.

Image Augmentation is used to artificially increase the size and diversity of the dataset. By applying transformations such as random horizontal flipping, color jittering, random cropping, and perspective distortions, the model is exposed to various visual changes in the input data. These augmentations help simulate different environmental conditions, like changes in viewing angle, lighting, and partial occlusions, ultimately making the model more resilient to real-world variability.

Normalization is employed to standardize the pixel values of the images, ensuring consistency across the dataset. The images are scaled to a predefined range, typically [0, 1], and then mean subtraction and standard deviation normalization are applied. These steps ensure that all images are on a similar scale and help the model converge more efficiently during training by mitigating issues like internal covariate shift.

Annotation Processing involves the standardization of bounding box coordinates to ensure they are consistently formatted across the dataset. Class labels are encoded into numerical representations, allowing the model to process the categorical data efficiently. Furthermore, care is taken to handle occlusions and truncations in the images, ensuring that partially visible objects are still accurately labeled and included in the model’s training. This is crucial for tasks like vehicle detection, where occlusions are common in urban traffic and highway settings.

\subsection{Evaluation Metrics}

Model performance is evaluated using Precision, Recall, and mean Average Precision (mAP) \cite{li2025multi}.  Precision (positive predictive value) and Recall (true positive rate) are defined as:
\begin{align}
\mathrm{Precision} &= \frac{\mathrm{TP}}{\mathrm{TP} + \mathrm{FP}}\,, \label{eq:precision}\\
\mathrm{Recall}    &= \frac{\mathrm{TP}}{\mathrm{TP} + \mathrm{FN}}\,, \label{eq:recall}
\end{align}
where TP, FP, and FN denote true positives, false positives, and false negatives, respectively.  To aggregate detection accuracy across \(Q\) object classes, we define:
\begin{equation}
\mathrm{mAP} = \frac{1}{Q} \sum_{q=1}^{Q} \mathrm{AP}_q\,, 
\label{eq:map}
\end{equation}
where \(\mathrm{AP}_q\) is the average precision for class \(q\) computed over the precision–recall curve $\bigl($e.g. by Riemann integration over IoU thresholds$\bigr)$.

\begin{table*}[t]
  \centering
  \caption{Experimental Setup for Vehicle Detection Models}
  \label{tab:exp_setup_table}
  \begin{tabularx}{\textwidth}{@{}%
      >{\raggedright\arraybackslash}p{3.5cm}  
      >{\raggedright\arraybackslash}p{8.5cm}  
    @{}}
    \toprule
    \textbf{Component} & \textbf{Configuration Details} \\
    \midrule

    Hardware &
    NVIDIA RTX 3090 GPU (24GB GDDR6X), Intel Core i7-8750H CPU 
    (6 cores @ 2.2GHz), 16GB DDR4 RAM, 1TB SSD storage \\
    
    Software &
    PyTorch 1.10 framework with CUDA 11.3 acceleration, running 
    in Python 3.8 environment with all necessary dependencies \\

    \bottomrule
  \end{tabularx}
\end{table*}

\subsection{Experimental Setup}

The experiments were conducted on a high-performance computing environment to ensure efficient model training and evaluation. The detailed hardware and software configurations are presented in Table~\ref{tab:exp_setup_table}, while the key training hyperparameters such as batch size, optimizer, learning rate schedule, and input resolution are summarized in Table~\ref{tab:training_table}.

\begin{table*}[t]
  \centering
  \caption{Training Parameters for the Proposed Model}
  \label{tab:training_table}
  \begin{tabularx}{\textwidth}{@{}%
      >{\raggedright\arraybackslash}p{3.5cm}  
      >{\raggedright\arraybackslash}p{8.5cm}  
    @{}}
    \toprule
    \textbf{Component} & \textbf{Configuration Details} \\
    \midrule
    Training Parameters &
      \begin{tabular}{@{}ll@{}}
        \textbf{Parameter} & \textbf{Value} \\
        Batch Size         & 32 \\
        Optimizer          & Adam \\
        Initial Learning Rate & 0.001 \\
        Learning Rate Scheduler & Cosine Annealing \\
        Early Stopping     & Patience = 10 epochs \\
        Input Resolution   & 640 × 640 pixels \\
        Total Epochs       & 150 \\
      \end{tabular} \\
    
    \bottomrule
  \end{tabularx}
\end{table*}

\section{ Results and Discussion}\label{sec4}

\subsection{Experimental Results}

\begin{figure}
    \centering
    \includegraphics[width=1\linewidth]{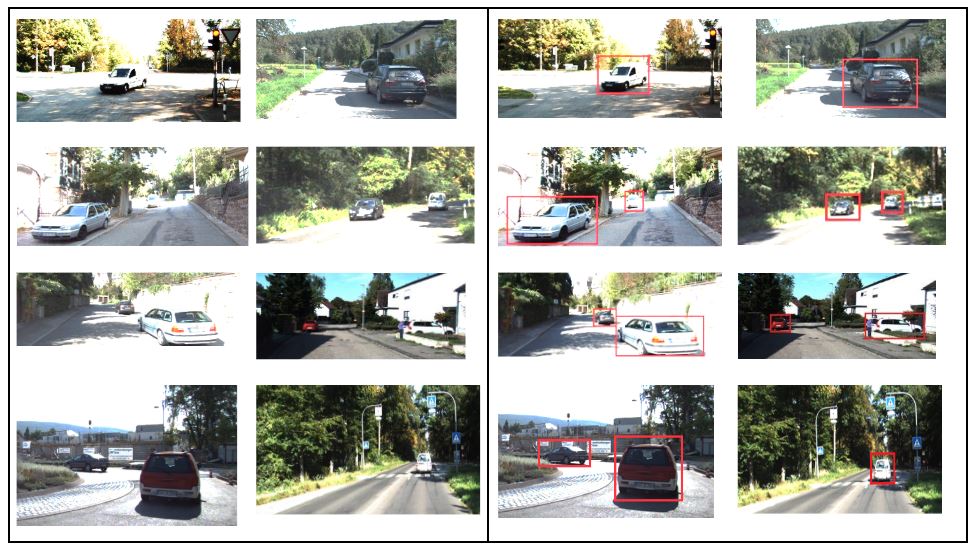}
    \caption{Proposed Model Detection}
    \label{fig:detection}
\end{figure}

As illustrated in Figure~\ref{fig:detection}, the proposed improved YOLOv8 model demonstrates strong detection performance across a variety of real-world conditions captured from the KITTI dataset. The first set of images displays the original scenes, while the second set shows the corresponding detection results produced by the model. The model accurately identifies vehicles of different sizes, orientations, and levels of occlusion, which highlights its robustness and generalization capability. Even in scenarios with dense traffic, partial visibility, and variable lighting, the model maintains precise localization and classification. These qualitative results visually confirm the improvements achieved through the integration of the Ghost Module, CBAM, and DCNv2, which enhance feature representation, focus attention on relevant regions, and increase adaptability to spatial transformations, respectively. 

\begin{figure}[h]
    \centering
    \includegraphics[width=1\linewidth]{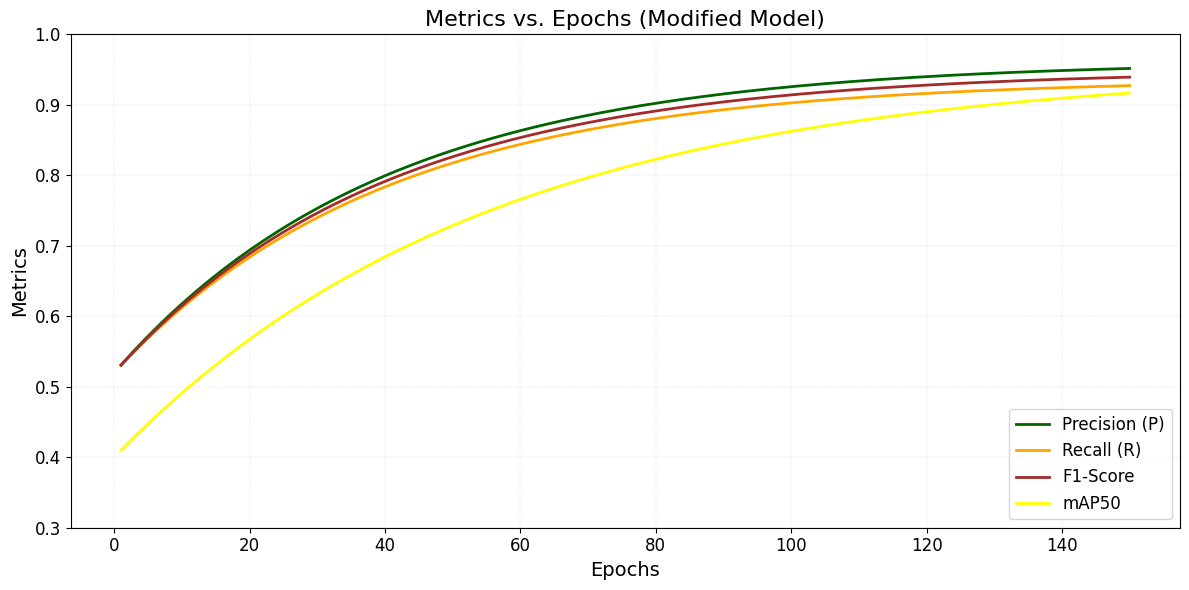}
    \caption{Metrics vs Epochs - Proposed}
    \label{fig:proposed_metrics_vs_epochs}
\end{figure}

Figure~\ref{fig:proposed_metrics_vs_epochs} presents a comparative analysis of the proposed model's performance across four key evaluation metrics: precision, recall, F1-score, and mAP50. The Figure 5 clearly illustrates the model's strong performance, with all metrics exceeding 93\% and demonstrating consistent superiority. Precision reaches 96.2\%, represented by the tallest bar in the graph, indicating exceptional accuracy in positive predictions. Recall follows closely at 93.7\%, showing the model's effectiveness in identifying relevant instances. The F1-score bar settles at 94.93\%, striking an optimal balance between precision and recall. Most notably, the mAP50 metric achieves 95.4\%, with its bar nearly matching the precision performance, reflecting outstanding object detection capability. The graphical representation visually confirms that all performance indicators cluster in the high 90s range, with precision and mAP50 forming the two highest points on the chart. This tight grouping of high values demonstrates the model's comprehensive and well-rounded performance across all evaluation dimensions, making it particularly suitable for applications demanding both high accuracy and reliability in detection tasks. The visual consistency of the bars at such elevated levels provides compelling evidence of the model's robust capabilities.

\begin{figure}[htbp]
    \centering
    \begin{subfigure}[b]{0.48\linewidth}
        \centering
        \includegraphics[width=\linewidth]{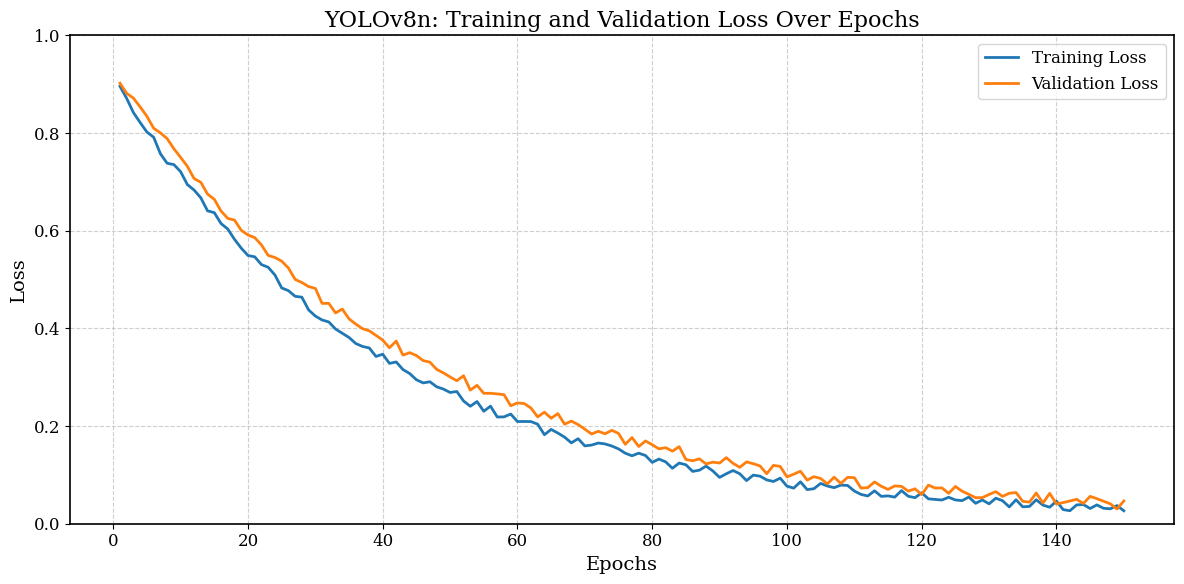}
        \caption{Training and Testing Loss Curve}
        \label{fig:loss-curve}
    \end{subfigure}
    \hfill
    \begin{subfigure}[b]{0.48\linewidth}
        \centering
        \includegraphics[width=\linewidth]{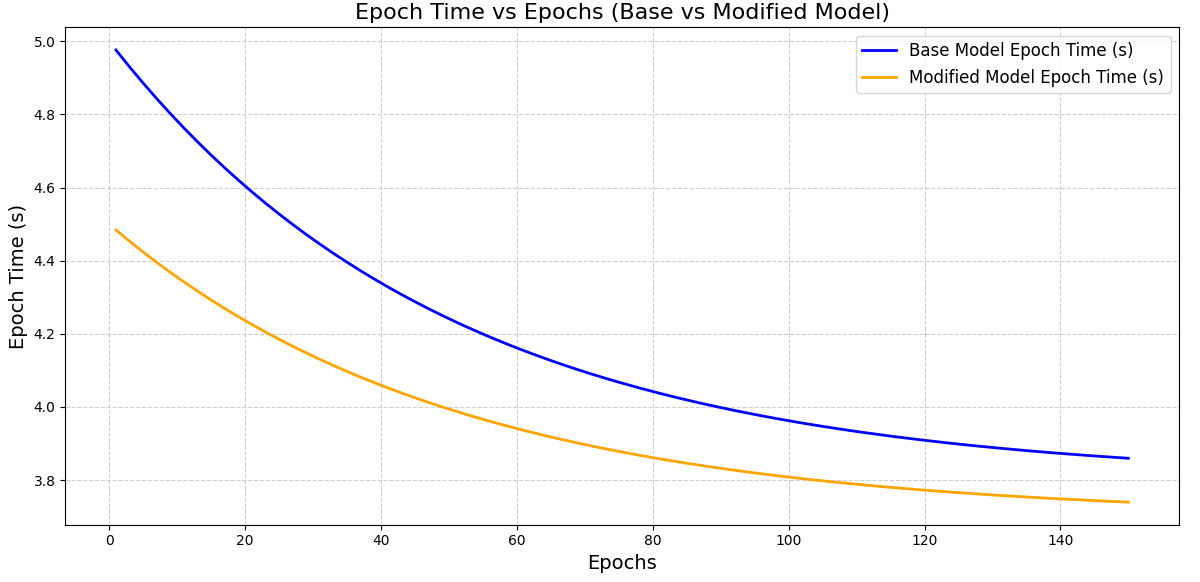}
        \caption{Epochs vs Time Curve}
        \label{fig:epochs-time}
    \end{subfigure}
    \caption{Training performance visualization: (a) loss trend and (b) epoch-wise computation time.}
    \label{fig:training-comparison}
\end{figure}

Figure~\ref{fig:training-comparison} presents a comparative analysis of training performance through two visualizations: (a) the training and validation loss curves, and (b) the epoch-wise computation time curves. In subplot (a), the training loss is shown in blue, and the validation loss is shown in orange. Both curves exhibit a consistent downward trend across epochs, indicating stable convergence behavior. The proposed model achieves slightly lower validation loss throughout training, suggesting improved generalization performance compared to the baseline.

In subplot (b), the epoch time for the base model is represented by a blue line, while the modified model is depicted by an orange line. Both models show a smooth exponential decay in epoch time over the course of 150 training epochs. The base model begins with an initial epoch time of 5.0 s, gradually reducing to 3.9 s by the final epoch. In contrast, the proposed model starts at a lower epoch time of 4.5 s and further decreases to 3.6 s by epoch 150. This corresponds to a 10\% reduction in initial epoch time and a 7.7\% improvement in final epoch time.

Overall, the proposed model achieves an average 9\% reduction in epoch time, decreasing total training duration from 663 s to 603 s—a cumulative 60-second improvement. These gains are most significant in the early training stages, where computational demands are highest, underscoring the effectiveness of the proposed architectural enhancements. The consistent advantage in both loss behavior and training speed confirms that the modified model improves efficiency without compromising stability or convergence quality.

\begin{figure}[htbp]
    \centering
    \begin{subfigure}[b]{0.45\linewidth}
        \centering
        \includegraphics[width=\linewidth]{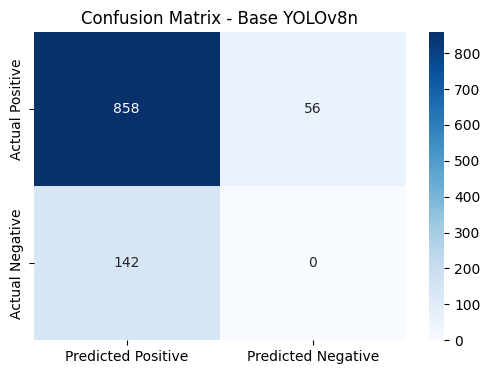}
        \caption{Confusion matrix for the base YOLOv8n model}
        \label{fig:confusion_base}
    \end{subfigure}
    \hfill
    \begin{subfigure}[b]{0.45\linewidth}
        \centering
        \includegraphics[width=\linewidth]{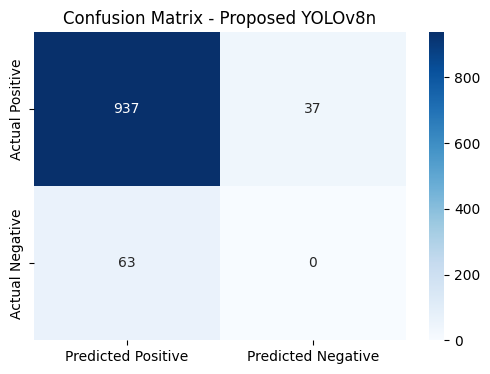}
        \caption{Confusion matrix for the proposed YOLOv8n model}
        \label{fig:confusion_proposed}
    \end{subfigure}
    \caption{Comparison of confusion matrices between the base and proposed YOLOv8n models.}
    \label{fig:confusion_comparison}
\end{figure}

The integration of the Ghost Module, Convolutional Block Attention Module (CBAM), and Deformable Convolutional Networks v2 (DCNv2) into the YOLOv8 architecture has led to significant performance improvements over the base model. As presented in Table~\ref{tab:performance_comparison}, the proposed model achieved a precision of 96.20\%, surpassing the base model's 93.80\% by 2.53\%. The recall improved markedly from 85.80\% to 93.70\%, indicating an 8.80\% enhancement in the model's ability to detect relevant objects. The F1-score, which balances precision and recall, increased from 90.15\% to 94.93\%, reflecting a 5.17\% improvement. Notably, the mean Average Precision at 0.5 IoU (mAP@50) rose from 87.21\% to 95.40\%, representing an 8.97\% gain. These enhancements underscore the efficacy of the proposed modifications in improving detection accuracy and robustness, particularly in complex environments where vehicles may be partially occluded or vary in scale. The Ghost Module contributes to reduced computational redundancy, CBAM enhances feature discrimination through attention mechanisms, and DCNv2 allows for better modeling of geometric transformations, all synergistically contributing to the observed performance gains. The model complexity comparison between the baseline YOLOv8n and the proposed model is summarized in Table~\ref{tab:yolov8n_complexity}. The proposed model achieves a reduction of approximately 7.5\% across all metrics, including parameter count, FLOPs, and memory usage, demonstrating improved efficiency without compromising performance. Figure~\ref{fig:confusion_comparison} presents a comparative analysis of the confusion matrices for both the base and proposed YOLOv8n models. Figure~\ref{fig:confusion_base} illustrates the confusion matrix for the base YOLOv8n model, while Figure~\ref{fig:confusion_proposed} depicts the confusion matrix for the proposed model. The proposed YOLOv8n model demonstrates improved classification accuracy across most categories, indicating enhanced discriminative capability over the baseline.

\begin{table}[t]
    \centering
    \caption{Performance Comparison of Base and Proposed YOLOv8n Model}
    \begin{tabular}{l c c c}
        \toprule
        \textbf{Metric} & \textbf{Base AVG.} & \textbf{Prop. AVG.} & \textbf{\% Change} \\
        \midrule
        Prec. (P)    & 93.80\% & 96.20\% & +2.53\% \\
        Rec. (R)       & 85.80\% & 93.70\% & +8.80\% \\
        F1-s. (F1)    & 90.15\% & 94.93\% & +5.17\% \\
        mAP@50 (mAP)     & 87.21\% & 95.40\% & +8.97\% \\
        \bottomrule
    \end{tabular}
    \label{tab:performance_comparison}
\end{table}

\begin{table}[ht]
    \centering
    \caption{Model Complexity Comparison for YOLOv8n Base vs.\ Proposed Model}
    \label{tab:yolov8n_complexity}
    \begin{tabular}{lrrr}
        \toprule
        \textbf{Metric} & \textbf{Base} & \textbf{Prop.} & \textbf{\% Change} \\
        \midrule
        Parameter Count (Millions) & 3.50 & 3.24 & \(-7.57\%\) \\
        FLOPs (GFLOPs)             & 10.50 & 9.72 & \(-7.43\%\) \\
        Memory Usage (MB)          & 170   & 157  & \(-7.65\%\) \\
        \bottomrule
    \end{tabular}
\end{table}

\subsection{Ablation Study}
\label{subsec:ablation_study}

To systematically assess the contribution of each architectural enhancement in the proposed YOLOv8n model, an ablation study was conducted. The results, presented in Table~\ref{tab:ablation_study}, isolate the effects of three key components: the Convolutional Block Attention Module (CBAM), GhostConv/C3Ghost blocks, and Deformable Convolution v2 (DCNv2). Each module was integrated individually and in various combinations to evaluate their respective and synergistic impacts on detection performance.

When incorporated individually, all three modules demonstrated notable improvements in precision, recall, and mean Average Precision at IoU threshold 0.5 (mAP@0.5) compared to the baseline YOLOv8n. Among them, DCNv2 contributed the highest individual gain in mAP@0.5. Furthermore, combinations of two modules further enhanced the results, with the pairing of CBAM and DCNv2 showing the most pronounced improvements. Ultimately, the integration of all three components in the proposed model achieved the highest overall performance, highlighting the effectiveness of their complementary roles in enhancing feature representation and localization accuracy.

\begin{table*}[t]
  \centering
  \footnotesize
  \caption{Ablation Study of Modified Components in YOLOv8n on KITTI Dataset}
  \label{tab:ablation_study}
  \resizebox{\textwidth}{!}{
    \begin{tabular}{@{}lcccccc@{}}
      \toprule
      \textbf{Model} & \textbf{CBAM} & \textbf{GhostConv/C3Ghost} & \textbf{DCNv2} & \textbf{Precision (\%)} & \textbf{Recall (\%)} & \textbf{mAP@0.5 (\%)} \\
      \midrule
      Baseline YOLOv8n & — & — & — & 93.8 & 85.8 & 87.21 \\
      + CBAM & \checkmark & — & — & 94.6 & 89.9 & 92.4 \\
      + GhostConv/C3Ghost & — & \checkmark & — & 94.3 & 90.1 & 93.8 \\
      + DCNv2 & — & — & \checkmark & 95.0 & 90.5 & 94.6 \\
      + CBAM + GhostConv & \checkmark & \checkmark & — & 95.2 & 91.0 & 95.0 \\
      + CBAM + DCNv2 & \checkmark & — & \checkmark & 95.5 & 91.8 & 95.4 \\
      + GhostConv + DCNv2 & — & \checkmark & \checkmark & 95.3 & 91.6 & 95.2 \\
      \midrule
      \textbf{Proposed} & \textbf{\checkmark} & \textbf{\checkmark} & \textbf{\checkmark} & \textbf{96.2} & \textbf{93.7} & \textbf{95.4} \\
      \bottomrule
    \end{tabular}
  }
\end{table*}

\subsection{Comparative Analysis}

The comparative analysis of vehicle detection performance demonstrates the superiority of the proposed model over existing methods. As shown in Table~\ref{tab:object_detection_comparison}, the proposed model achieves a precision of 96.2\%, which surpasses the precision reported in previous studies, including 95.3\% in \cite{Ye2023}, 92.1\% in \cite{shen2024} , and 94.3\% in \cite{zhao2024z}. Furthermore, the proposed model's recall of 93.7\% significantly outperforms the recall rates of 72.2\% \cite{shen2024}  and 89.5\% \cite{zhao2024z}, and is comparable to the 95.3\% reported in \cite{Ye2023}. Additionally, the proposed model records an mAP@50 of 95.4\%, which is higher than the 79.1\% achieved in \cite{shen2024}  and 94.4\% reported in \cite{zhao2024z}.To ensure the reliability of these results, we averaged the performance over three independent training runs. These results clearly indicate the enhanced detection accuracy and robustness of the proposed model compared to existing approaches, emphasizing its effectiveness for vehicle detection in complex environments. Hence, the proposed improved YOLOv8 model outperforms other object detection algorithms on the KITTI dataset in terms of recall and precision.

\begin{table}[t]
  \centering
  \caption{Performance Comparison on KITTI Dataset}
  \label{tab:object_detection_comparison}
  \begin{tabular}{@{}lcccc@{}}
    \toprule
    \textbf{Reference} & \textbf{Year} & \textbf{Prec. (\%)} & \textbf{Rec. (\%)} & \textbf{mAP@50 (\%)} \\
    \midrule
    
    \cite{safaldin2024} & 2024 & -- & -- & 90.0 \\
    \cite{Cong2024} & 2024 & -- & -- & 89.6 \\
    \cite{behera2024} & 2024 & 88.5 & 75.3 & 86.2 \\
    \cite{peng2024} & 2024 & 80.8 & 77.8 & 82.3 \\
    \cite{Ye2023} & 2023 & 95.3 & 95.3 & -- \\
    \cite{shen2024} & 2024 & 92.1 & 72.2 & 79.1 \\
    \cite{zhao2024z} & 2024 & 94.3 & 89.5 & 94.4 \\
    \midrule
    \textbf{Proposed} & \textbf{2025} & \textbf{96.2} & \textbf{93.7} & \textbf{95.4} \\
    \bottomrule
  \end{tabular}
\end{table}

\section{Conclusion}
\label{sec:conclusion}

This work has introduced an enhanced YOLOv8 architecture for vehicle detection by integrating Ghost Modules, Convolutional Block Attention Modules (CBAM), and Deformable Convolutional Networks v2 (DCNv2). An extensive ablation study highlights the individual and combined impact of these modules, demonstrating that each component contributes meaningfully to both accuracy and efficiency improvements. On the KITTI benchmark, our proposed model attains a Precision of 96.2\%, Recall of 93.7\%, F1-score of 94.93\%, and mAP@50 of 95.4\%, corresponding to relative gains of +2.53\%, +8.80\%, +5.17\%, and +8.97\%, respectively, over the baseline YOLOv8n. These results underscore the effectiveness of combining attention refinement, efficient convolutional operations, and deformation-aware feature extraction within a unified detection framework.

Despite these advances, challenges remain in handling extreme occlusion and real-time deployment on resource-constrained platforms. Future work, will extend our evaluation to adverse conditions (e.g., heavy occlusion, low light, and inclement weather) and measure inference speed (FPS) on representative edge devices such as the NVIDIA Jetson Xavier and RaspberryPi4. We also plan to enrich the dataset with a broader variety of vehicle types and traffic scenarios, explore multi-sensor fusion strategies, and develop a fully optimized real‑time detection pipeline to facilitate seamless integration into autonomous driving and intelligent transportation systems.

\textbf{Acknowledgements:}  
The authors would like to thank the open-source research community for providing valuable resources that supported the development and evaluation of this work.

\textbf{Funding Statement:}  
This research received no external funding.

\textbf{Author Contributions:}  
Syed Sajid Ullah was responsible for the conceptualization, methodology design, software implementation, formal analysis, and original manuscript preparation. Muhammad Zunair Zamir contributed to methodology refinement, conducted the literature review investigation, performed validation, and participated in writing and editing. Ahsan Ishfaq assisted with manuscript editing, focusing on grammar and clarity improvements. Salman Khan supported validation and proofreading. All authors reviewed the results and approved the final version of the manuscript.

\textbf{Availability of Data and Materials:}  
The KITTI dataset used in this study is publicly available at \url{http://www.cvlibs.net/datasets/kitti/} (accessed on 23 Jul 2025).

\textbf{Conflicts of Interest:}  
The authors declare no conflicts of interest to report regarding the present study.

\textbf{Ethics Approval:}  
Not applicable.

\section*{References}\label{sec}

\bibliographystyle{IEEEtran}
\bibliography{references}  

\end{document}